\documentclass{article}
\usepackage{ijcai18}
\usepackage[keeplastbox]{flushend}
\usepackage{graphicx}
\usepackage[export]{adjustbox}
\usepackage{amsmath}
\usepackage{algorithm}
\usepackage[noend]{algpseudocode}
\usepackage{balance}
\usepackage{tcolorbox}
\usepackage{listings}
\usepackage{setspace}
\usepackage{times}
\usepackage{xcolor}
\usepackage{soul}
\usepackage[utf8]{inputenc}
\usepackage[small]{caption}

\usepackage{graphicx,subfigure}
\algnewcommand{\LineComment}[1]{\State \(\triangleright\) #1}

\newcommand{\tuple}[1]{\mbox{$\langle #1 \rangle$}}

\newcommand{\haz}{\cal H\,}

\newcommand{\A}{\mbox{$\cal A\,$}}
\newcommand{\E}{\mbox{$\cal E\,$}}

\newtheorem{theorem}{Theorem}[section]

\newtheorem{corollary}[theorem]{Corollary}

\newenvironment{definition}[1][Definition]{\begin{trivlist}
\item[\hskip \labelsep {\bfseries #1}]}{\end{trivlist}}

\newcommand{\RNum}[1]{\uppercase\expandafter{\romannumeral #1\relax}}

\makeatletter
\patchcmd{\@verbatim}{\verbatim@font}{\verbatim@font\small}{}{}
\makeatother

\usepackage{wrapfig}
\usepackage{caption}

\title{Antifragility for Intelligent Autonomous Systems}
\author{
Anusha Mujumdar, 
Swarup Kumar Mohalik,
Ramamurthy Badrinath
\\ 
Ericsson Research, Bangalore \\
{anusha.pradeep.mujumdar, swarup.kumar.mohalik,ramamurthy.badrinath}@ericsson.com
}

\begin{document}
\maketitle
\begin{abstract}
Antifragile systems grow measurably better in the presence of hazards. This is in contrast to fragile systems which break down in the presence of hazards, robust systems that tolerate hazards up to a certain degree, and resilient systems that -- like self-healing systems -- revert to their earlier expected behavior after a period of convalescence. The notion of antifragility was introduced by Taleb for economics systems, but its applicability has been illustrated in biological and engineering domains as well. In this paper, we propose an architecture that imparts antifragility to intelligent autonomous systems,
specifically those that are goal-driven and based on AI-planning. We argue that this architecture allows
the system to self-improve by uncovering new capabilities obtained either through the hazards themselves (opportunistic) or through deliberation (strategic). An AI planning-based case study of an autonomous wheeled robot is presented. We show that with the proposed architecture, the robot develops antifragile behaviour with respect to an oil spill hazard. 
\end{abstract}

\section{Introduction}

How should an intelligent, autonomous system adapt to unforeseen situations? Answering this question has been the focus of significant research effort over the last two decades \cite{hayes1995architecture}. However, the notion of adaptivity has so far been limited to a system’s capability to cope with unknown situations and return, at best to its original performance. In this work, we argue that the presence of unforeseen stressors, or {\em hazards}, offers a system an opportunity to self-improve. Such self-improvement in the presence of hazards is the central notion of {\em{antifragility}}.

The term antifragility was coined by Nicholas Taleb in his book \cite{taleb2012antifragile}, in which he argued that the opposites of {\em fragile} systems are not {\em robust} or {\em resilient} systems, but systems that benefit from hazards and grow stronger as a result. These concepts and their relationships are compelling and have been illustrated in many domains such as 
economics, biology and engineering. However, the literature does not provide any suggestion towards formalization of the concepts, due to which there are no concrete guidelines
for designing an antifragile system. In this paper, we address this gap by formalizing the notions of fragility, robustness, resilience and antifragility in the context of intelligent systems, in particular, for a large subclass of systems that are based upon knowledge-based reasoning and AI planning. We show that 
antifragile intelligent systems can be designed using a refinement of MAPE-K (Monitor, Analyse, Planning and Execution - Knowledge), the autonomic architecture suggested by IBM ~\cite{IBMmapek}.


The paper is organized as follows. We briefly review  some background work on antifragility in section \ref{sec:background}. Section \ref{sec:scope} lays out the scope of the current study, and describes our approach towards introducing antifragility concepts into intelligent systems. Section \ref{sec:sys_model} details the intelligent systems model followed, and formalizes concepts of fragility and antifragilility in such systems. Section \ref{sec:implementation} proposes design modules for antifragility, to be augmented to an existing autonomic computing architecture of intelligent systems. In section \ref{sec:example}, we illustrate the developed concepts with the help of a robot path planning example. We conclude in section \ref{sec:conclusion} with some future directions.

\section{Related Work}
\label{sec:background}
Systems that degrade in behavior when acted upon by external hazards (or, stressors) are known as {\em fragile} systems.
{\em Robust} systems display tolerance to hazards upto a certain pre-designed level, and show no change in performance. If the magnitude of the hazard exceeds this level, system performance rapidly degrades. An example of a robust system is a building designed to withstand an earthquake of a particular magnitude. {\em Resilient} systems are affected by hazards, but return to normal performance after a hiatus. Robust and resilient systems have been studied extensively as a means to allow intelligent systems to cope with hazards~\cite{russell2015research}; however, in such systems, the performance, at best, retains its original behavior. 

Taleb notes that certain systems, in fact, benefit from the presence of hazards, and that these systems need to be distinguished from being merely robust or resilient. Such systems abound in nature e.g., the immune system strengthens itself upon encounter with an antigen and often retains the developed coping capability permanently~\cite{monperrus2017principles}. Another example is muscular strength being developed by the stressor of exercise~\cite{Jones2014}. The increase in strength so developed improves the capability of the muscle to perform a wide variety of tasks many of which may be unrelated to the exercise that initially triggered it. In environmental ecosystems \cite{taleb2012antifragile}, the presence of a hazard in the form of a predator strengthens the surviving prey population, since they learn from past experience.


Antifragile systems have been studied in the literature in the context of communication systems \cite{Lichtman2016a}, infrastructure networks \cite{Fang2017}, aerospace \cite{Jones2014}, health care ~\cite{Clancy2015} and cloud systems \cite{Johnson2013}. All of these emphasize the following characteristics of antifragile systems:
\begin{itemize}
\item Such systems thrive in the presence of hazards, as opposed to robust systems that simply endure, and resilient systems which merely survive and return to original performance at best.
\item Antifragile systems learn from failure to handle hazards; the failure may be their own or of other members in a population-based system.
\item When acted upon by hazards, the system responds in a fashion that improves its performance over time (not necessarily immediately); the improvement is through acquisition of new capabilities caused by deliberate strategies by the system or through the opportunistic exploitation of certain attributes of the hazard itself.
\end{itemize}
The literature, in general, does not provide formal definitions of the concepts of fragility, robustness, resilience and antifragility.
In  \cite{Lichtman2016a}, antifragility is demonstrated mathematically for a particular hazard in a communication system,
but the definitions are specific to the problem and are not generic enough to be applicable to other systems. 

\section{Our Scope and Approach}
\label{sec:scope}
In this paper, we wish to formalize the concepts of antifragility in the context of intelligent systems. Specifically, we
consider the class of systems that are goal-driven and based on AI planning. These systems have a fixed set of actions. Given goals and constraints, they attempt to achieve the goals by deriving a sequence (called {\em plan}) from the available actions and then executing the plan. 

An intelligent system can be fragile because hazards encountered during execution may drive the system to states from where it cannot find a plan to achieve its goals. If, on the other hand, it is able to find an alternate plan to lead to the goals no matter where the hazards drive it to, it is designated resilient. Note that if the system is able to find plans that avoid the hazards altogether then occurrence of hazards do not affect the plans - such a system is robust w.r.t. the hazards. This is possible in many cases since most hazards are localized (e.g. traffic jam, corruption of memory block).


A crucial point to note is that robustness and resilience uses plans over actions available with the system. However, antifragile systems, by definition, must become stronger when a hazard occurs. In the intelligent systems, we correlate {\em strength} with {\em available plans} and take a position that a system is stronger if there are more plans available. This can be implemented by introducing new actions in the system. As a consequence, a system that was fragile w.r.t. some hazards may now become robust and/or resilient. This is consistent with the observation in antifragility literature that new capabilities introduce redundancy, which brings beneficial effects through resilience and robustness.

Modeling the above notion of antifragility in the context of intelligent systems has a difficulty. In the antifragility literature, new capabilities are introduced through external entities (e.g. medicine) or through {\em realization} of already existing capabilities (e.g. exercise).
However, the intelligent systems that we consider have fixed sets of actions and
do not have access to any external source for new actions\footnote{We note that the generalization to ``open" intelligent systems is certainly interesting, and we leave this direction for future work.}. 
We ameliorate this situation by partitioning the set of actions into visible and hidden subsets.
At any point of time, the plan synthesis algorithms have access to only the visible subset. 
Actions can be brought in from the hidden set to introduce new actions for the
planner. One can argue that such an approach is ``artificial" because the system by default has access
to all actions, which results in the largest set of plans possible. However, 
our approach is realistic and has a strong rationale: first, due to the smaller visible set, planning is more efficient. More importantly, there is a cost to support the planning and execution of actions (more sensors and actuators, more predicates, larger state space). Therefore, it is advisable to consider only a subset of the actions that are necessary in the current situation and enhance the subset only when in need (e.g. to handle a hazard). This is accurately captured by the partition of the actions into visible and hidden sets.


An antifragile intelligent system must therefore first decide what actions are made visible, when and for what duration, and then execute these decisions. We show that by suitable refinement of the MAPE-K loop mentioned in the introduction, with refined action sets and appropriate modules, one can obtain an architecture to design intelligent systems that exhibit antifragility. In the rest of this section, we give some details of the base MAPE-K loop.

\begin{figure}[h]
	\centering
	\includegraphics[width=60mm]{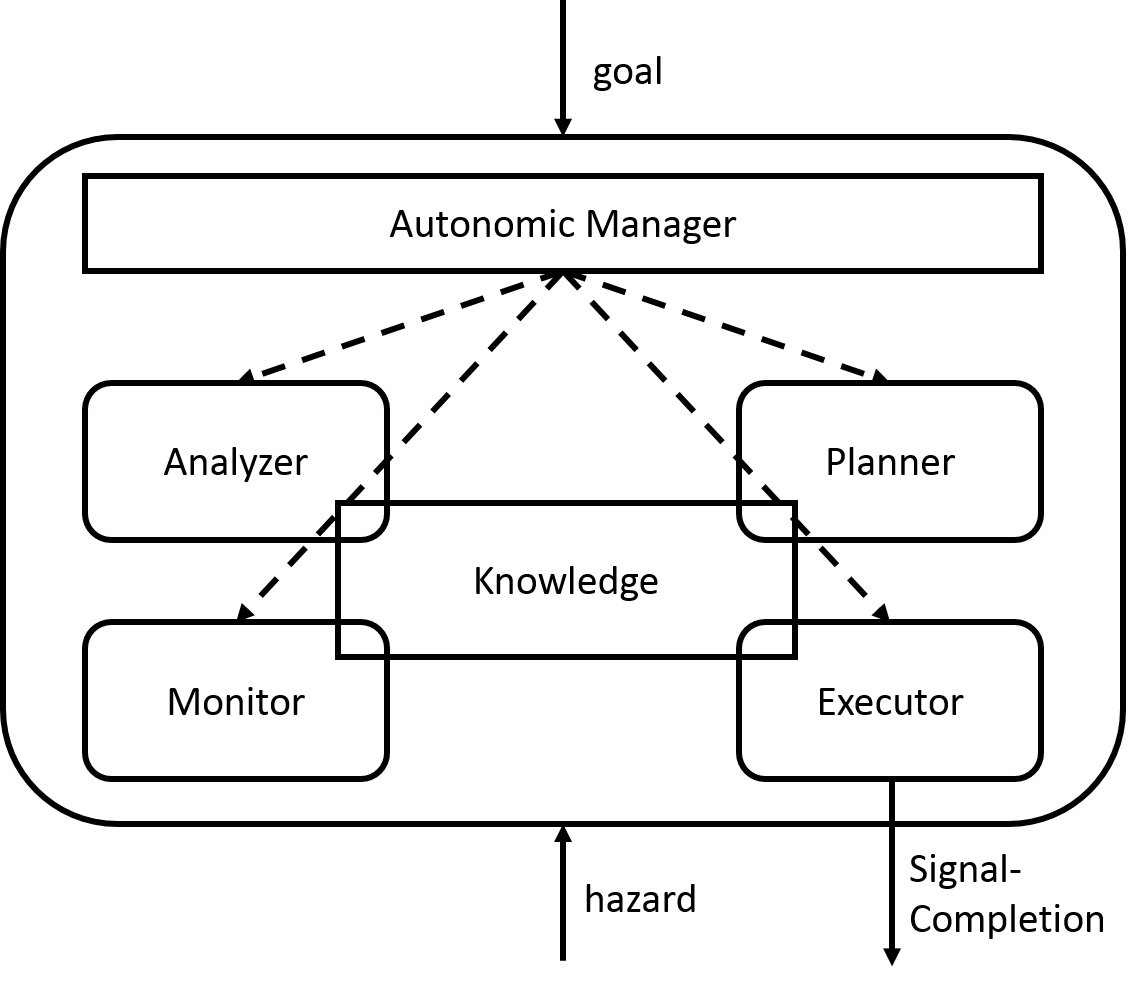}
	\caption{Intelligent System with MAPE-K Architecture}
	\label{fig:mape-k}
\end{figure}

Intelligent systems based on AI planning can be implemented using the MAPE-K loop (see Fig.~\ref{fig:mape-k}). It comprises the Monitor, Analyzer, Planner, Executor and Knowledge modules orchestrated in a loop by the Autonomic Manager. The Monitor module is responsible for gathering relevant state information from the resources to detect the events of interest. This information is then used by the Analysis 
module to determine if there is a need to change the current plan (sequence of actions) which is being 
executed. The change may be necessitated due to a change in the goals of the system or 
changes in the environment which force a course correction. The Planning module is then called upon to 
synthesize a new plan for the changed goals/context. The Execution module uses the generated plan  
to resume its execution. During the monitoring of the environment and execution, the system derives knowledge about 
the environment and possible responses that it can store in a knowledge-base . This knowledge is used to derive faster and better responses as the system continues to interact with the environment.

\section{System Model}
\label{sec:sys_model}

We model an intelligent system \A as a goal-driven, planning-based agent interacting with an 
environment \E. The interaction is specified as a game where the environment supplies a goal 
and the agent tries to achieve the goal through a plan (sequence of actions). Once a goal is 
achieved, the environment supplies another goal and thus the game continues ad infinitum.
During the execution of the plan, the environment may change the state of the system 
nondeterministically, capturing the notion of hazards. The type of response from \A determines 
whether it is fragile, robust, resilient or antifragile. 

Formally, we have a finite set of boolean predicates $Q$, and the system states $S$ are defined 
as all valuations $V:Q \rightarrow \lbrace ${\it true, false}$ \rbrace$. The environment \E is specified through a set of goal 
states $G \subseteq S$, goal transitions  $T_G \subseteq G\times G$ (also called missions)
and a set of hazard transitions $H \subseteq S\times S$.  For a hazard $(s, t)$, $s$ is referred 
to as the {\em hazard source} and $t$ the {\em hazard consequence}. 
There is a special subset $W$ of $G$ called the {\em waypoints}.  
%

The intelligent system \A is defined over of a set of actions $Act$ where each action is specified by a 
\tuple{precondition, effect} pair with {\it precondition}, {\it effect} $\subseteq Q$. $Act$ induces a deterministic
labeled graph $S_{Act}$ with $S$ as the nodes and edges $(s, a, t)$ when $precondition(a)$ is consistent 
with $s$ and $t = s\leftarrow$ {\it effect}$(a)$ denoting the values of the predicates in $s$ being overridden by {\it effect}$(a)$.
A sequence of actions $P$, is a {\em plan} for a pair of states $(c, g)$ if starting at the 
initial state $c$ and applying the actions in $P$ consecutively, we arrive at the goal state $g$. The corresponding
path is denoted as $path(P, c, g)$. We assume that the system has a special {\em reset} action which moves it to one 
of the waypoints i.e. for every state $s \in S$, there is a waypoint $w \in W$ such that $(s,~ reset,~ w) \in S_{Act}$.

\begin{figure}[h]
	\centering
	\includegraphics[width=60mm]{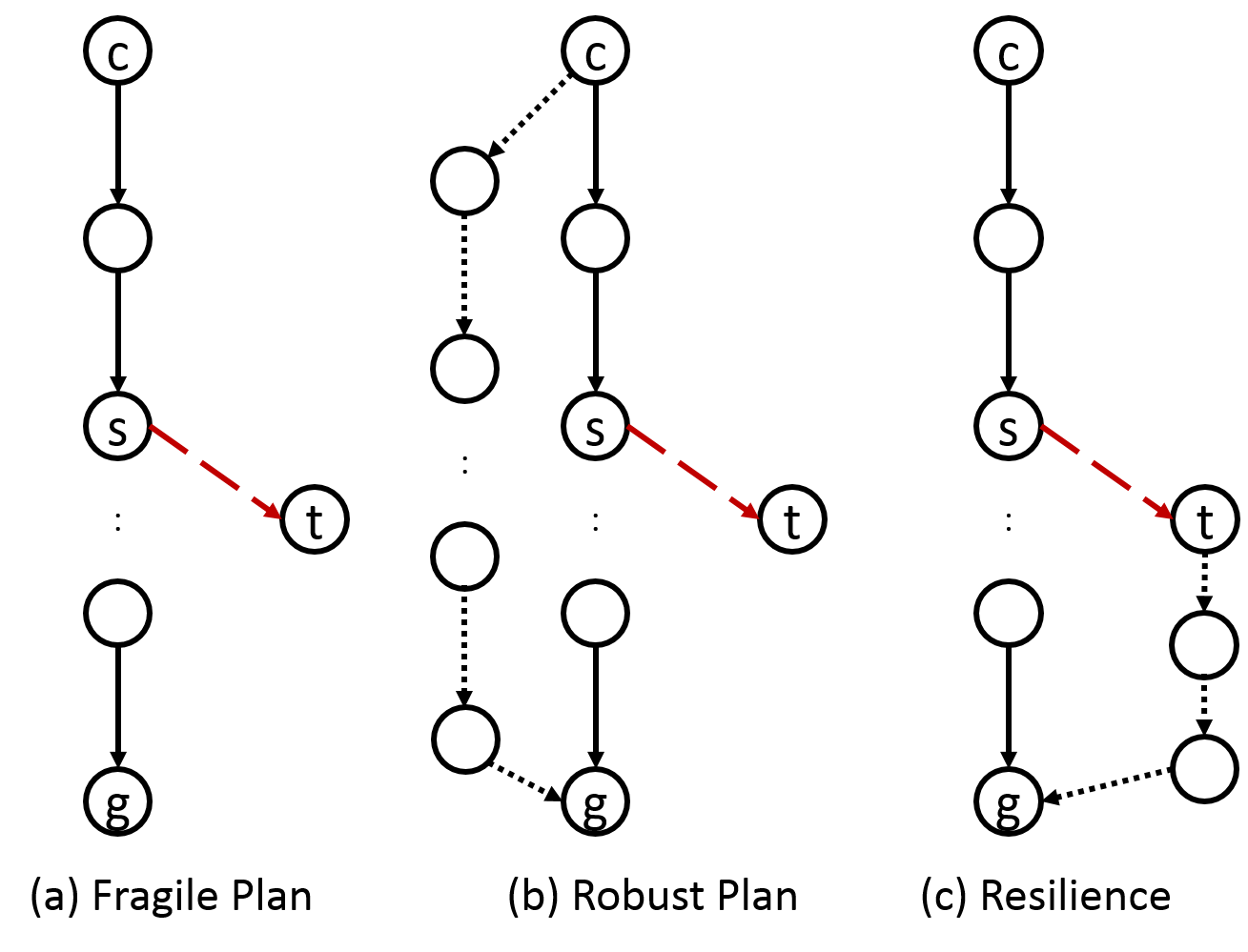}
	\caption{Fragile, Robust and Resilient Plans}
	\label{fig:type-plans}
\end{figure}

The minimal notations above are sufficient to define the following concepts succinctly.
\begin{definition}
A plan $P$ for $(c, g)$ is {\em fragile} w.r.t $H$ if there is state $s$ on $path(P, c, g)$ and $(s,t) \in H$ such that there is no plan $P'$ for $(t, g)$.

A plan $P$ for $(c, g)$ is {\em robust} w.r.t a set of hazards $H$, if for all $(s, t) \in H$,  $s$ is not on $path(P, c, g)$. 

A plan $P$ for $(c, g)$ is {\em resilient} w.r.t $H$, if for all $(s, t) \in H$, $s$ is on $path(P, c, g)$ implies there is a plan $P'$ for $(t, g)$.
\end{definition}
A fragile plan is one which goes through a state where a hazard can occur and the from the hazard consequence state one cannot achieve the intended goal state. 
A robust plan completely avoids the states where hazards can occur. A resilient plan switches to another starting from the hazard consequence state when a hazard occurs. These plans are illustrated in Figure~\ref{fig:type-plans} (in (b) the dotted path on the left corresponds to a robust plan). One can extend these definitions to the entire system.

\begin{definition}

An intelligent system is {\em fragile} w.r.t $H$ if there is a mission $(c, g)$ such that all plans for $(c, g)$ are fragile w.r.t. $H$.

An intelligent system is {\em robust} w.r.t a set of hazards $H$ if for every mission $(c, g)$, there is a robust plan w.r.t. $H$.

An intelligent system is {\em resilient} w.r.t. $H$ if, for every mission $(c, g)$, any plan $P$ for $(c, g)$ is resilient w.r.t. $H$.
\end{definition}

\begin{corollary}
An intelligent system is not fragile w.r.t. $H$ if it is either robust or resilient w.r.t. $H$.
\end{corollary}
Robustness and resilience are independent concepts, i.e. a system may be robust w.r.t. a hazard $h$ because there is a robust plan for every mission, but there may be other
plans which enable a hazard and are not resilient. Similarly, all the plans for the missions may be resilient but there may not be a single robust plan.

Fragility, robustness and resilience refers to a given system. However, antifragility points to the capability of systems to evolve from {\em fragile} to {\em not fragile}.
This is how we define the notion of ``an antifragile system becomes stronger" due to the occurrence of hazards.

\begin{definition}
An intelligent system \A is {\em Antifragile} w.r.t. a set of hazards $H$ if, 
\A is fragile w.r.t $H$ implies \A eventually becomes either resilient or robust to $H$.
\end{definition}
A corollary of the definition is that whereas there was no plan for $(t, g)$ for some hazard consequence $t$ and goal $g$ originally, there needs to be such plan to have resiliency. Similarly, there has to be new paths for some mission $(s, g)$ to have robustness. This implies there is a need to introduce new actions for the agent. 

As one can see, the definition of antifragility leaves a number of dimensions unspecified. When there is a hazard, if there is no alternate plan, it does not specify whether the improvement through new actions should happen immediately or at a future time, whether the improvement should be for a limited duration (say, till the current goal) or permanent, whether the improvement should be for the specific hazard only or a related larger set. We note that such decisions depend upon predictive capabilities of the system and cost considerations that must be left for the designer. 


\section{Implementation}
\label{sec:implementation}
In this section, we describe the refinements for the MAPE-K loop and also suggest some design decisions for the unspecified dimensions mentioned in the previous section as a guide for system designers.


\subsection{Knowledge}
The Knowledge module (KM) has a set of actions. As observed in Section~\ref{sec:scope}, the key idea for introducing new actions is partitioning the actions $Act$ into 3 classes: $Act_e$(empowering), $Act_v$(visible) and $Act_h$(hidden). All actions $a$ have an associated visibility predicate $visible_a$. The membership of action $a$ in $Act_v$ or $Act_h$ is determined by the boolean value of $visible_a$ in the current state. Visibility predicates can be set to True only through empowering actions, or by the hazards themselves, though they can be set to False by all actions and hazards. 

Apart from the actions, the knowledge base maintains a history of hazards that have occurred and a set of visibility predicates called {\em internal goals}. The latter are essentially the predicates which can possibly be achieved through planning and execution of empowering actions. Through this, the Manager can make some of the hidden actions visible in the future in a deliberate and strategic way.

\subsection{Monitor}
The normal role of Monitor in MAPE-K is to access the current state of the system and detect anomalies if any. Here, the Monitor detects a hazard through the inconsistency of the current state ($c$) and the desired precondition of the action-to-be-executed. In case of a hazard, it records the hazard $(e, c)$ in the knowledge base, where $e$ is the state that had been reached after the execution of the previous action.

\subsection{Analyzer}
The design decisions we mentioned are mostly in the Analyzer module. The Analyzer  refers to the history of the hazards, domain knowledge including cost etc (which could be based upon analytics, for example, and taking into account the impact of the hazard on the achievement of the current goal) and outputs the following in the event of a hazard:

\begin{enumerate}
\item It first determines a set of hazards (including the present one) that needs to be handled, by predicting and/or correlating with the present hazard. 
\item It outputs for each hazard a three-tuple \tuple{W, D, M} where W denotes immediate or future improvement (\{Now, Later\}), D denoted duration of improvement (\{Current, All\}) and M denotes mode of handling(\{Robust, Resilient\})
\end{enumerate}
Note that the \tuple{W, D, M}'s are only recommendations. The system will ``improve" for any value of M. W denotes that the urgency of improvement and D controls whether the new actions are necessary for longer term.

\subsection{Planner}
Given the current state, a goal state, set of actions, a set of hazards and the recommendation from the analyzer, the Planner synthesizes a plan using only the visible actions. First, it determines the current state space from the predicates used in $Act_v$. Note that since $Act_v$ is dynamic, the state space also changes dynamically. It is expected that the Monitor can query the environment and find the correct current state at any level of detail (e.g. accessing the appropriate sensors and processing modules outputting the predicates).

Plan synthesis may fail because of several reasons. All of visible or hidden actions may not be sufficient to provide new plans. This is a pathological case where external support is mandatory. There may not be plans satisfying the recommendation in which case the best effort improvement is done. For example, if a robust plan cannot be found, the Planner tries to find a resilient one. Or, if robust plan cannot be found for all missions, it finds one for the current mission and so on. Specific behavior depends upon Planner policies.


\subsection{Executor}
The Executor module is pretty simple: given a plan which is a labeled sequence of transitions in $S_{act}$,
it executes the actions associated with the labels in the sequence order taking help of the Monitor to detect hazards. If there is a hazard, the Executor halts
the execution, signals the Manager that a hazard has taken place and appends the hazard to the history in the knowledge base. In the best case, the Executor finishes executing the plan without countering
any hazard and signals the environment about plan completion. The Environment then can issue a new goal
which is handled by Autonomic Manager.

When the Manager decides to issue a {\em reset} action, it is issued directly to the Executor. Just as 
it executes a plan, the Executor executes the reset action which leads to a waypoint (a special goal state).
The reset completion signal to the Environment allows it to continue the game with a new goal. 

\subsection{Autonomic Manager}
Autonomic manager is essentially an orchestrator for other components in MAPE-K, but
has a greater role in antifragile intelligent systems since the life cycle is no longer just a loop.






The reactive behavior of the Manager is as follows:
\begin{enumerate}
\item When \E issues a new goal, Manager passes the goal along with the $Act_v$, corresponding current state, hazards and \tuple{W, D, M} recommendations to the Planner for plan synthesis.

\item When \E introduces a hazard, the Manager gets the signal from the Executor. 
It invokes the Planner as in (1) to first find if there is a resilient plan already in
the current $S_{Act_v}$. If no such plan exists, then it invokes the Analyzer to get possibly a set of hazards $H'$ with the \tuple{W, D, M} recommendations. For space reasons, assume that there is a single hazard to be handled. The case of multiple hazards is routine though tedious.
\begin{enumerate}
\item
Manager invokes Planner to use both visible and hidden actions to come up with a plan $P$ with minimum number of hidden actions if possible depending upon mode M of the hazard $h$. If it is possible to synthesize a plan, then it records the visibility predicates $Pred_v(h)$ of the hidden actions in $P$. 
\item
If W==Later, then Manager issues reset to the Executor, but in a separate, parallel thread triggers the Planner to achieve the visibility predicates and execute the plan subsequently. This is so that during the planning and execution for other missions, hidden actions are enabled gradually.
\item
If W==Now, then Manager triggers the planner to first achieve $Pred_v(h)$, which
ensures new actions in $Act_v$ and then invoking $P$ for the current goal.
\item
If D == Current, then the visibility predicates are toggled by the Manager as soon as the current goal is achieved.
\end{enumerate}






\end{enumerate}
Thus, whenever there is a hazard, if there is a possibility of having a robust or resilient system w.r.t. the hazard using extra hidden actions, those actions are made visible. This ensures that whenever there is a new goal, the system is able to find 
a plan (robust or resilient) for the goal. Note that for completely different goals,
the system may not find robust/resilient plans in $S_{Act}$, hence a similar exercise is carried out to make more actions visible.

In the antifragility literature, it is noted that certain systems may perform better under hazards (e.g. adrenaline rush enabling great strength in stress conditions). In a way, this attribute is captured in our setup by the fact that a hazard may lead to a state where certain visibility predicates are set to True and hence the agent has access to (possibly) more number of visible actions for planning. On the other hand, building up resources to enable more strength or immunity is a deliberate process and this is captured by the internal planning with empowering actions.

\begin{figure*}
\centering     
\subfigure[Originally planned path]{\label{fig:a}\includegraphics[width=50mm]{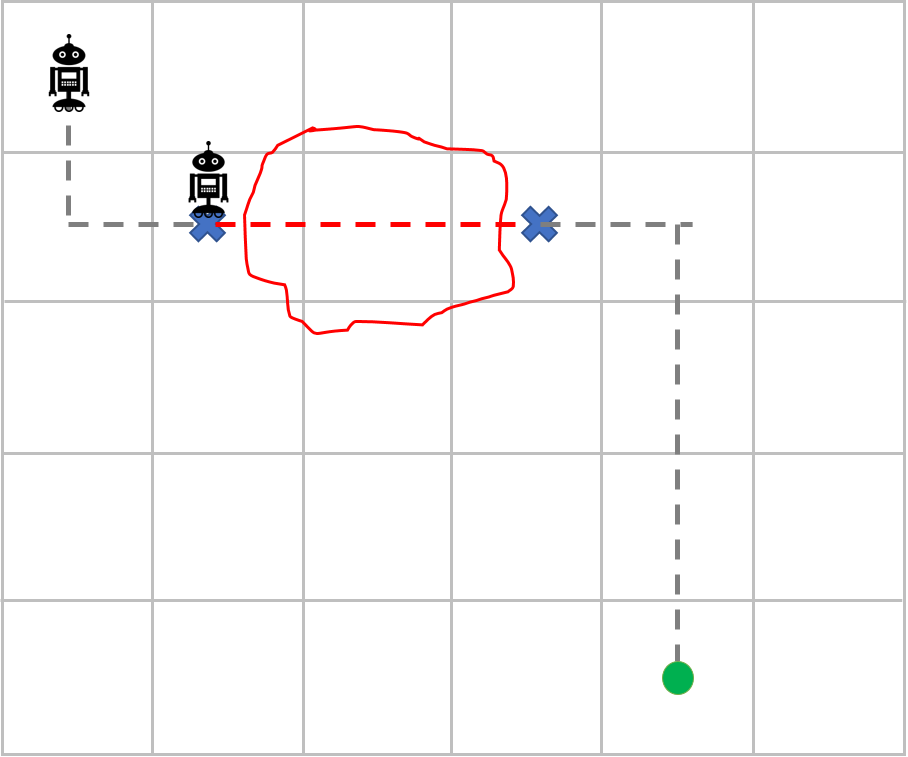}}
\hspace{0.5cm}
\subfigure[Hazard mode triggered by oil-spill hazard]{\label{fig:b}\includegraphics[width=86mm]{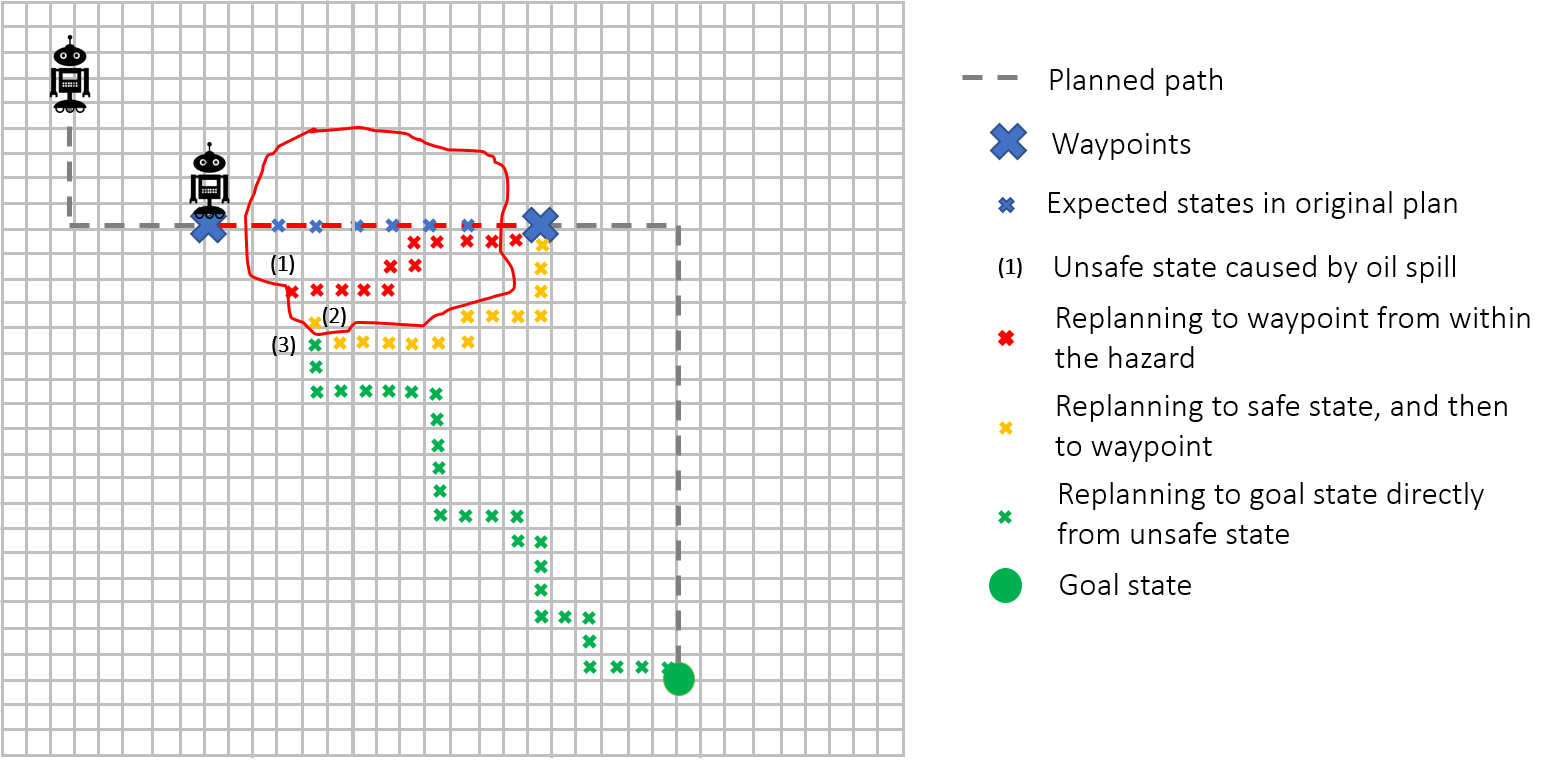}}
\caption{Robot path planning example: first goal sequence}
\end{figure*}
\begin{figure}[h]
	\centering
	\includegraphics[width=50mm]{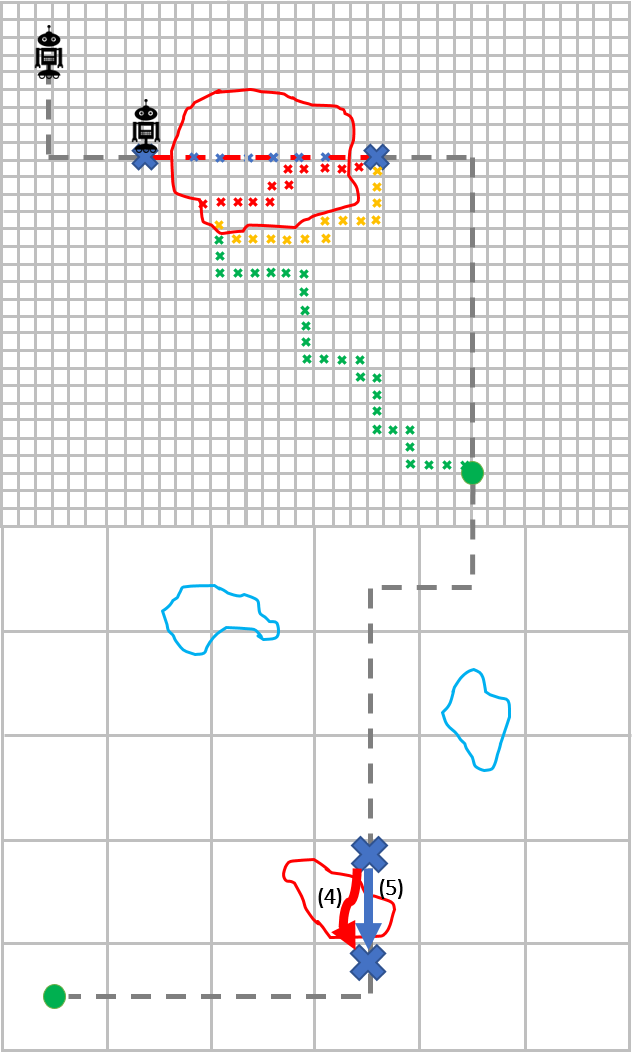}
	\caption{Robot path planning example: second goal sequence}
	\label{fig:haz_future}
\end{figure}

\section{Case Study}
\label{sec:example}
We illustrate the concepts discussed in this paper within a path planning scenario. We consider a wheeled robot \A autonomously navigating in its world (e.g., a warehouse). The mission is split into a sequence of goals, with the next goal  being planned for once the current goal is successfully achieved. Each goal is nothing but a position the robot has to reach on a sparse 5$\times$6 grid of equally sized cells.

In a simplistic scenario, the robot has the following visible actions: MOVE and TURN using which it navigates the sparse grid. Traditional path planning is used to compute the optimal sequence of actions that take the robot from an initial position to the given goal. The robot's behavior also has two other actions \textit{small}MOVE and \textit{small}TURN. These two are initially not visible (hidden actions).  These two require additional support from the environment during execution, for example they need sensors to locate the robot on a finer grid. Therefore we assume additional \textit{empowering} actions are typically needed before one can use these hidden actions; the preconditions for these actions account for that requirement through appropriate predicates.

During execution, \A follows the computed plan. An oil spill, unknown at the time of planning, is encountered, which throws the robot off the planned path and hence the original plan is not applicable (shown by point (1) in Figure \ref{fig:b}). Thus the robot discovers it has encountered a hazard. 

The Autonomic Manager now uses external information (camera images, image analytics etc.) to identify that the hazard is an oil spill and uses domain knowledge to derive the fact that MOVE and TURN actions cannot be used in the oil spill. Therefore it makes these actions Hidden, and sets the visibility predicates of {\it small}MOVE and {\it small}TURN as an internal goal. The Planner uses the prerequisite empowering actions to prepare the current state and ultimately makes the {\it small}MOVE and {\it small}TURN actions visible. The robot's domain view when these actions are enabled is shown in Figure \ref{fig:b}. To support this world view, note that the environment needs to be able to provide the right kind of sensory information (e.g., fine grid positions).

For the purposes of this illustration, we assume that {\it small}MOVE and {\it small}TURN actions are deterministic and have the intended effect\footnote{A reasonable assumption when referring to sufficiently small steps on slippery surfaces.}. Various possibilities exist for replanning with these additional actions and it depends upon the Planner.
As shown in Figure \ref{fig:b}, the robot can navigate from within the oil spill, with {\it small}MOVE and {\it small}TURN actions to a known state on the earlier plan (red path), it can come out of the oil spill as early as possible and then navigate the the known state (yellow path) and it could directly find the a path to the current goal, abandoning the earlier plan. As we have mentioned, getting out of the oil spill may be treated as a hazard as a response to which MOVE and TURN actions be made visible, and the robot now has all the four actions at its disposal for planning the new paths.

Figure \ref{fig:haz_future} shows a second goal sequence after the first goal has been successfully reached. We note that the path avoids known oil spills, due to the awareness of this new entity in the environment. While following the path, a new oil spill that did not exist at plan-time may be encountered. In this case, the robot uses its newly visible actions to easily cope. In Figure \ref{fig:haz_future}, the red curve (4) shows the path using these new capabilities. We also note that these new actions may also be used as planned actions to navigate \textit{through} known oil spills even at original plan time possibly resulting in a lower cost path. Thus in reaching the second goal if the red oil spill was known, it could still have chosen the path through the oil spill, the blue line - (5), using the newly visible actions for this stretch appropriately.

As mentioned in the Implementation, Analyzer with the help of Knowledge module can generalize the specific oil spill  hazard
to other hazards such as ice, slippery sand etc. by identifying that the hazard consequence state is similar and also
from the knowledge that these surfaces have similar properties. With this, similar uncovering of hidden actions can be
done for all such hazards. This shows the dimension of the antifragile architecture where occurrence of one hazard
builds antifragility for a class of other hazards.

This illustrative example has shown that it is possible for an intelligent system to develop antifragility through the suggested refinements and suitable design decisions in the base MAPE-K architecture.

\section{Summary and Further Work}
\label{sec:conclusion}

In this paper we have defined antifragility in the context of intelligent  systems which are goal-directed and based on AI planning. We have shown that antifragile intelligent systems can be built using the MAPE-K architecture though with several additional features. Primarily, the available action set should be divided into empowering, visible, and hidden subsets. Depending on the strategy of the Autonomic Manager module, the system is made robust or resilient to the hazard in different manners, correctly capturing the essence of antifragility. The proposed architecture has been illustrated through a robotics path planning example, where a robotic agent \A performs better after encountering a hazard.

One immediate problem arises that must be studied: that of optimization of the selection process through which hidden actions are made visible. Efficiency of algorithms to select the minimal number of new actions that will allow robustness/resilience is critical from the view of real-time systems. 

We observe that the improvement of the systems is triggered by hazards in antifragility literature. In an indirect way, by defining hazards as any violation of cost metrics, one can bring in improvements through new actions that will result in more efficient plans even though the system does not encounter ``unforeseen" hazards. The {\em simian army} in~\cite{Abid2014} used to
inject faults and create hazards to improve the systems is similar in spirit.

We note that this work is an initial effort to study the concept of antifragility in the context of intelligent systems, and therefore is rather preliminary. Undoubtedly, rigorous implementation and verification of the developed ideas is needed. In addition, the notion of performance improvement from antifragility, albeit intuitively true, must be quantified in the current context. We believe that the real value of building antifragile systems will become evident as we develop complex multi-agent systems. Therefore, distributed antifragility, built into a population of systems, is an interesting future direction. The potential applications of a system that improves from the presence of hazards is vast. We hope that the promise of antifragile intelligent systems will stimulate further research within the intelligent systems community.

\bibliographystyle{named}
\bibliography{citations}
\end{document}